\documentclass[10pt,twocolumn,letterpaper]{article}

\usepackage{cvpr}              

\usepackage[dvipsnames]{xcolor}

\definecolor{cvprblue}{rgb}{0.21,0.49,0.74}
\usepackage[pagebackref,breaklinks,colorlinks,citecolor=cvprblue]{hyperref}
\usepackage{times}
\usepackage{epsfig}
\usepackage{graphicx}
\usepackage{amsmath}
\usepackage[linesnumbered,ruled,vlined]{algorithm2e}
\usepackage{amssymb}
\usepackage{array}
\usepackage{booktabs}
\usepackage{amssymb}
\usepackage{pifont}
\usepackage{comment}
\usepackage{caption}
\usepackage[accsupp]{axessibility}
\usepackage[numbers]{natbib}

\SetCommentSty{commentfont}

\newcommand{\DAVID}[1]{{\textcolor{cyan}{David: #1}}}

\newcolumntype{?}{!{\vrule width 1.2pt}}

\newcommand{\OURS}{UnScene3D}
\def\paperID{1852} 
\def\confName{CVPR}
\def\confYear{2024}

\title{\OURS{}: Unsupervised 3D Instance Segmentation for Indoor Scenes}

\author{
David Rozenberszki$^{1}$~~~~~~
Or Litany$^{2,3}$~~~~~~
Angela Dai$^1$
\vspace{0.2cm} \\ 
$^1$Technical University of Munich~~~
$^2$Technion~~~
$^3$NVIDIA
\vspace{0.2cm} \\ 
\href{https://rozdavid.github.io/unscene3d}{https://rozdavid.github.io/unscene3d}
}

\begin{document}

\twocolumn[{%
\renewcommand\twocolumn[1][]{#1}%
\maketitle
\vspace{-1cm}
\begin{center}
    \centering
    \captionsetup{type=figure}
    \includegraphics[width=\textwidth,keepaspectratio]{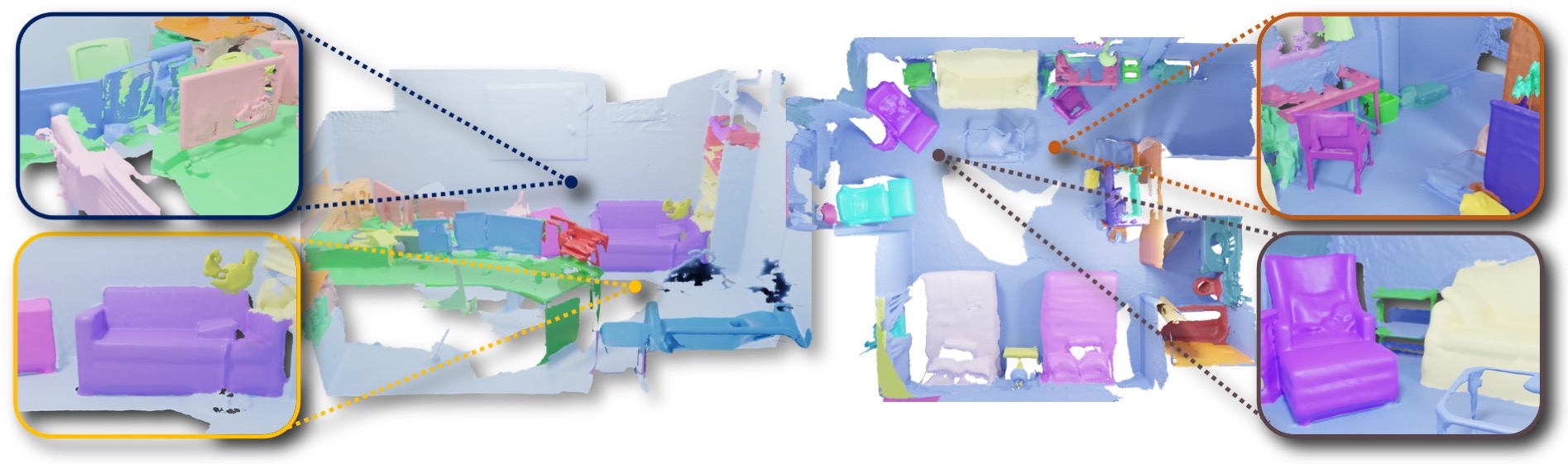}
    \vspace{-0.7cm}
    \captionof{figure}{We propose \OURS{}, a fully-unsupervised 3D instance segmentation method, effectively separating semantic instances without requiring any manual annotations. 
    We utilize geometric primitives to ensure crisp masks, and due to our self-training loop, we can also obtain a dense set of predictions, even in cluttered indoor scenarios.
    }
\end{center}%
}]

\maketitle
\begin{abstract}
3D instance segmentation is fundamental to geometric understanding of the world around us. 
Existing methods for instance segmentation of 3D scenes rely on supervision from expensive, manual 3D annotations.
We propose \OURS{}, the first fully unsupervised 3D learning approach for class-agnostic 3D instance segmentation of indoor scans.
\OURS{} first generates pseudo masks by leveraging self-supervised color and geometry features to find potential object regions.
We operate on a basis of geometric oversegmentation, enabling efficient representation and learning on high-resolution 3D data.
The coarse proposals are then refined through self-training our model on its predictions.
Our approach improves over clustering-based alternatives to unsupervised 3D instance segmentation methods by more than 300\% Average Precision score, demonstrating effective instance segmentation even in challenging, cluttered 3D scenes.
\end{abstract}
\section{Introduction}

The increasing availability of commodity RGB-D sensors, now widely available on iPhones as well as with the Microsoft Kinect or Intel RealSense, has enabled consumer-level capture of 3D geometry of real-world environments.
To enable applications in robotics, autonomous navigation, and mixed reality in such scenes, semantic 3D scene understanding is necessary. In particular, 3D instance segmentation is critical to 3D perception, providing dense instance mask predictions, thus enabling physical and geometric reasoning about objects in an environment.
While various 3D deep learning approaches have been developed for 3D instance segmentation \cite{qi2017pointnet++,wu2019pointconv,wang2019dynamic,rethage2018fully,wang2018sgpn,hu2020randla,fan2021scf,occuseg, chen2021hierarchical,hou20193d,rozenberszki2022language,liang2021instance,vu2022softgroup,vu2022softgroup++,hui2022graphcut,kolodiazhnyi2023topdown,sun2022spformer,Schult23mask3d}, they require full supervision from expensive, manual, dense annotations on 3D scenes.

We introduce \OURS{}, a novel approach designed for class-agnostic 3D instance segmentation. 
Our aim is to identify objects in real-world 3D scans by predicting their dense instance masks, without any constraints to a predefined set of class categories. 
Moreover, we avoid expensive data annotation requirements by operating in an unsupervised fashion, instead leveraging self-supervised 2D and 3D features for segmentation.


\OURS{} comprises two essential elements.
First, we observe that for RGB-D scan data, self-supervised representation learning methods \cite{xie2020pointcontrast,hou2021exploring} can provide an innate signal indicating object-ness through feature similarity. 
We thus generate pseudo masks over 3D segment primitives, based on multimodal analysis of self-supervised color and geometry features from the RGB-D data.
By considering mesh segments rather than voxels or points, our approach efficiently scales with high-resolution 3D data in large scene environments while inherently promoting contiguous segmentation masks. 
As we require strong features for these initial coarse estimates, we fuse information from both geometric and 2D color features in a complementary fashion.
Second, following the pseudo mask generation, we train our model through iterative self-training on both the initial pseudo masks and the current confident model predictions. Through multiple rounds of self-training with noise robust losses achieve improved object recognition and segmentation.
At inference time, we do not require any 2D color signal and can produce class-agnostic 3D instance segmentation for a new geometric observation of a 3D environment.
Experiments on challenging, cluttered indoor environments from the ScanNet~\cite{dai2017scannet}, S3DIS \cite{armeni_s3dis} and ARKit~\cite{dehghan2021arkitscenes} datasets show that \OURS{} improves significantly over unsupervised, clustering-based state of the art. 
%
In summary, our contributions are:
\begin{itemize}
	\item We propose an unsupervised 3D instance segmentation approach for indoor RGB-D scans, without requiring any human annotation.
	\item We generate sparse 3D pseudo masks for unsupervised training based on a multi-modal fusion of color and geometric signal from RGB-D scan data. We achieve robustness and efficiency through a geometry-aware scene coarsening. 
	\item Our generated pseudo masks are iteratively refined by self-training for 3D instances to improve 3D instance segmentation performance.
\end{itemize}

\section{Related Work}

\paragraph{Self-supervised 3D pretraining}

While significant progress has been made in fully supervised 3D instance segmentation \cite{qi2017pointnet++,fan2021scf,3DSemanticSegmentationWithSubmanifoldSparseConvNet,choy20194d,occuseg,hou20193d,rozenberszki2022language,vu2022softgroup++,sun2022spformer,hou2023mask3d} the amount of densely annotated 3D data is scarce. Inspired by success in the 2D domain, various 3D pretraining methods have been developed to boost semantic and instance segmentation performance when fine-tuning with annotated semantic labels. 
Such methods leverage instance discrimination based on different camera views \cite{xie2020pointcontrast,hou2021exploring}, local augmentations \cite{zhang_depth_contrast}, or multiple LIDAR sweeps \cite{nunes2022segcontrast}.
While these methods can provide powerful 3D feature extraction, they do not construct any notion of object instances.

\paragraph{Weakly-supervised 3D segmentation}
Classical methods have leveraged object template information to match or retrieve templates to local geometry in a scene \cite{searchclassify,kim2012acquiring,karpathy2013object,chen2014automatic,li2015database,Nakajima_2019_ICCV}, thereby identifying potential object locations.
Other methods formulated 3D dense instance segmentation with only 3D box annotation \cite{chibane2021box2mask, Peng2023PointCI} or single-point supervision and active-learning \cite{Liu2021OneTO, Wang2022OneCO}. 
More recent methods have focused on exploiting knowledge from powerful pre-trained vision-language models to inform text-guided queries in 3D scenes \cite{shafiullah2022clipfields, conceptfusion, Peng2022OpenScene, ding2022language, Liu2022PartSLIPLP}; however, such methods still rely on large-scale annotated data in the 2D domain.

\paragraph{Clustering-based segmentation}
There has been very little work done in fully unsupervised 3D instance segmentation, but classical clustering methods have been used to group regions with similar geometric properties together. 
A particularly notable approach is the density-based clustering of DBSCAN~\cite{dbscan} and its hierarchical counterpart HDBSCAN~\cite{mcinnes2017accelerated_hdbscan}. 
These methods can be used to group point clusters in a 3D scene based on point normals and colors.
The ScanNet dataset~\cite{dai2017scannet} showed that the Felzenswalb algorithm~\cite{felzenszwalb2004efficient} originally developed for image over-segmentation, can generate useful geometric segment clusters. 
We also exploit such geometric primitives to guide dimensionality reduction and feature aggregation. 

Finally, recent methods have been developed to detect instances with self-supervised pretrained features in driving scenarios. 
These methods often leverage the unique properties of such data  including dynamics and instance sparsity. 
Song et. al.~\cite{song2022_ogc} identify object instances through  motion, showing promise for self-driving scenarios, but limited to moving objects.
Nunes et. al.~\cite{nunes2022unsupervised} additionally propose a clustering and graph cut based refinement on pre-trained 3D features, focusing on sparse outdoor scenarios to identify spatially separate objects. 
Our solution aims to segments instances  in complex, cluttered indoor environments. 

\paragraph{Unsupervised 2D instance segmentation}

Classical graph-cut algorithms \cite{wu1993_mincut,Chopra1993ThePartitionProblem,Deza2009GeometryOC,shi2000normalized_cut} can be used to detect objects in scenes, employing low-level feature clustering to identify self-similar regions.
Recent advances in self-supervised feature learning have been employed in  2D unsupervised instance segmentation methods, which  use two-stage training pipelines to achieve remarkable segmentation results \cite{wang2022freesolo,wang2023cut}. 
These methods first generate a set of coarse pseudo masks building on the insights of graph-cut algorithms and then refine them with a series of self-training iterations. 
In particular, FreeSolo~\cite{wang2022freesolo} uses multi-branch feature extraction to obtain self-similar regions as mask proposals, producing a dense set of initial pseudo-annotated instances.
CutLER~\cite{wang2023cut} uses the normalized cut (NCut) algorithm~\cite{shi2000normalized_cut} with deep self-supervised features from DINO~\cite{caron2021emerging_dino} to identify multiple prominent regions as pseudo masks.  
%
Inspired by such approaches we also leverage pseudo mask generation and self-training, but to handle high-dimensional, noisy real-world 3D scan data, we employ a multi-modal feature reasoning and geometric graph coarsening for robust unsupervised 3D instance segmentation. 
\section{Method}\label{sec:method}
\begin{figure*}
    \centering
    \includegraphics[width=\linewidth,keepaspectratio]{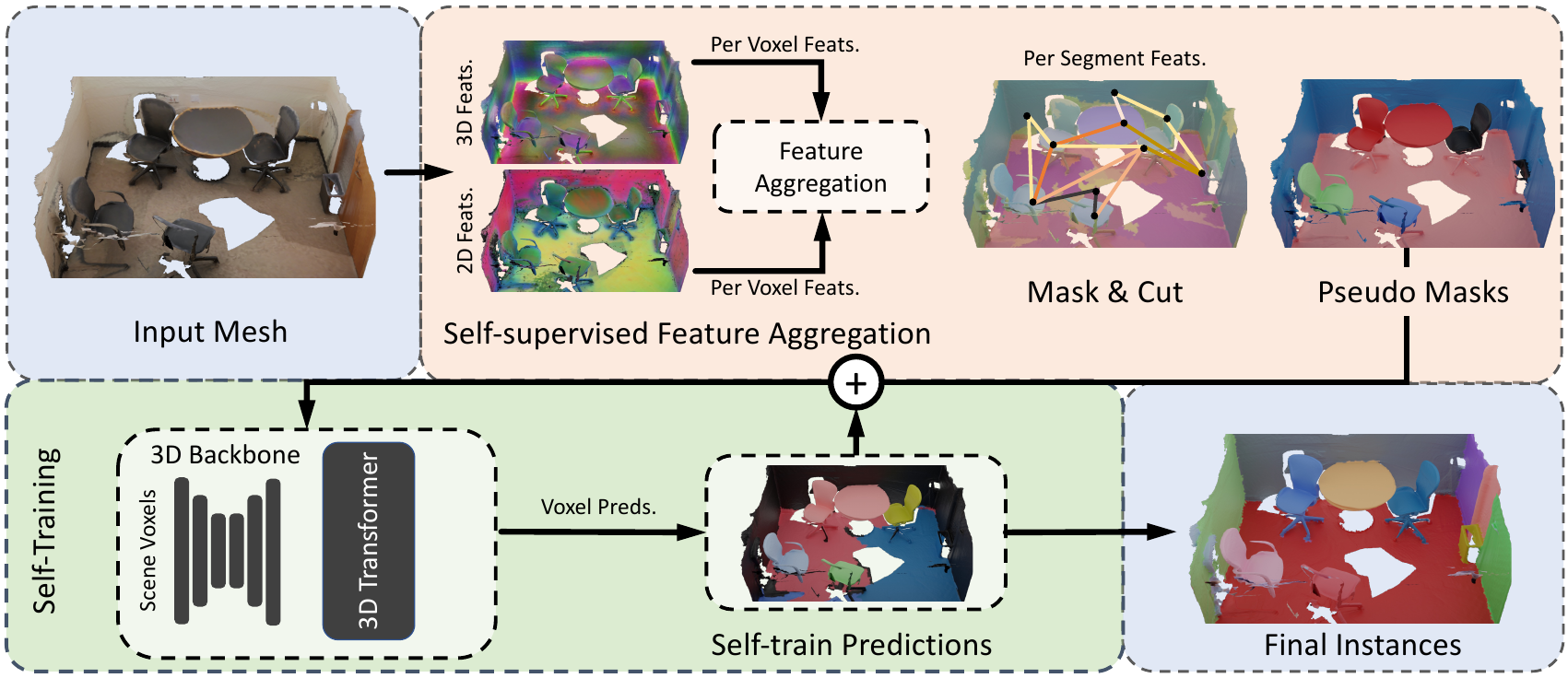}
    \vspace{-0.7cm}
    \caption[]{
    \OURS{} first generates a set of pseudo masks (top) to initiate self-training (bottom) for unsupervised 3D instance segmentation.
    We leverage features from 3D self-supervised pre-training in combination with 2D self-supervised features on an input mesh.
    These multi-modal features are then aggregated on geometric primitives, integrating low- and high-level signals for pseudo mask segmentation.
    These initial pseudo masks are then used as supervision for a 3D transformer-based model to produce updated instance masks that are integrated into the supervision of multiple self-training cycles.
    Finally, we obtain clean and dense instance segmentation without using any manual annotations.} 
    \label{fig:full_pipeline}
    \vspace{-0.4cm}
\end{figure*}

\paragraph{Problem definition}
We propose an unsupervised learning-based method for 3D instance segmentation.
We operate on a set of training 3D scenes $\{X_i\}_{i=1}^{n_t}$, represented as mesh graphs $G = (V, E)$, of vertices $V$ and triangular face edges $E$, where each scene $X_i$ contains an unknown set of $n_i$ objects in the $i^{th}$ scene. We aim to train a model that can predict for a previously unseen input scene $X$, a set of 3D masks representing the different object instances in that scene. 

\vspace{-0.3cm}
\paragraph{Method overview}
In order to achieve unsupervised 3D instance segmentation we first break down the scenes into $N$ geometric primitives $S_N$, which we use to initialize an adjacency matrix $W$ to extract an initial set of pseudo masks $M^0$, representing instance hypotheses based on combining 2D and 3D inputs $\mathcal{F}_{2D}$ / $\mathcal{F}_{3D}$ $\in R^{N \times D_{2D/3D}}$, where $D_{2D}, D_{3D}$ are the dimensions of the $2D/3D$ self-supervised features. We regularize the per-segment similarities over geometric primitives for mitigating noise and enabling efficient 3D reasoning.
We then employ a series of self-training cycles, updating pseudo mask supervision with new predicted masks, in order to produce final 3D instances.
An overview of our approach is shown in Figure~\ref{fig:full_pipeline}.

\subsection{Initial pseudo mask generation}\label{sec:dataset_gen}

In order to initiate self-training, we first generate an initial set of pseudo masks, leveraging complementary information from 2D and 3D signal in $\{X_i\}$.

\subsubsection{Feature aggregation}

To encourage effective initial pseudo mask generation, we employ joint reasoning across both self-supervised color and geometry features, as they can provide complementary information regarding objects.
As RGB-D scans often contain color image information and reconstructed 3D scan geometry, we can associate both 2D and 3D features in 3D by back-projecting the 2D extracted features using the corresponding depth and camera pose information for each image. 
Both 2D and 3D features are extracted through state-of-the-art self-supervised feature learning methods \cite{hou2021exploring, caron2021emerging_dino}.
As real-world camera estimation often contains small misalignment errors and noise or oversmoothing in reconstructed scan geometry, these self-supervised features can often also contain high-frequency noise, which we address in Sec.~\ref{sec:ncut_primitives} when reasoning over these features.
Note that while we employ both 2D and 3D signal when available for training, we do not require any aligned color image inputs for inference, enabling more general applicability.

\subsubsection{3D Graph Cut}\label{sec:ncut_primitives}
To generate pseudo masks from the 2D and 3D self-supervised features, we employ graph cut to estimate class-agnostic instances from the background. 
More precisely, we leverage the principle of Normalized Cut~\cite{shi2000normalized_cut} (NCut), which employs eigenvalue decomposition from an adjacency matrix $W\in R^{N \times N}$ over a graph to identify self-similar regions potentially representing semantic instances, where a set of potential instances can be extracted iteratively. 
Given a graph representing the 3D scene, we build an adjacency matrix $W$ and self-supervised features with a corresponding degree matrix $D\in R^{N \times N}$, where $D(i,i) = \Sigma_jW(i,j)$ and $(D-W)v = \lambda D v$. 
In this system, finding the second smallest eigenvalue $\lambda$  and its corresponding eigenvector $v$ is a close approximation for the minimized cost.
From $v$, we obtain foreground separation by taking all node activations where the eigenvector components were larger than their mean.
To identify multiple foreground objects, this process is repeated iteratively.

Unfortunately, applying this approach directly to the 3D scenes $\{X_i\}$ in common 3D representations such as voxels or points is not only computationally infeasible, but unreliable due to the noise in camera pose estimation and geometric reconstruction of 3D scan data. 
Thus, we propose to regularize the graph cut across geometric primitives.


\subsubsection{Geometric Primitives}


To employ efficient reasoning across high-dimensional 3D data and enable robust 3D regularization of noisy features, we propose to operate on geometric primitives acquired through a graph coarsening process.
For a 3D scene $X_i$ we construct the graph $G = (V, E)$ where $V$ and $E$ being the mesh vertices and face edges. Then, nodes with similar normals and colors are aggregated and clustered based on the mesh topology following \cite{felzenszwalb2004efficient} and resulting in a set $S_N = \{C_1 \dots C_N \}$ and $\bigcup (S_N) = V$ where $C_n$ represent a single primitive. 
This reduces the graph size by multiple orders of magnitude, and enables effective regularization of noise in the used self-supervised 2D and 3D features.

\subsubsection{NCut on Geometric Primitives}

\begin{figure}[!ht]
    \centering
    \includegraphics[width=\linewidth,keepaspectratio]{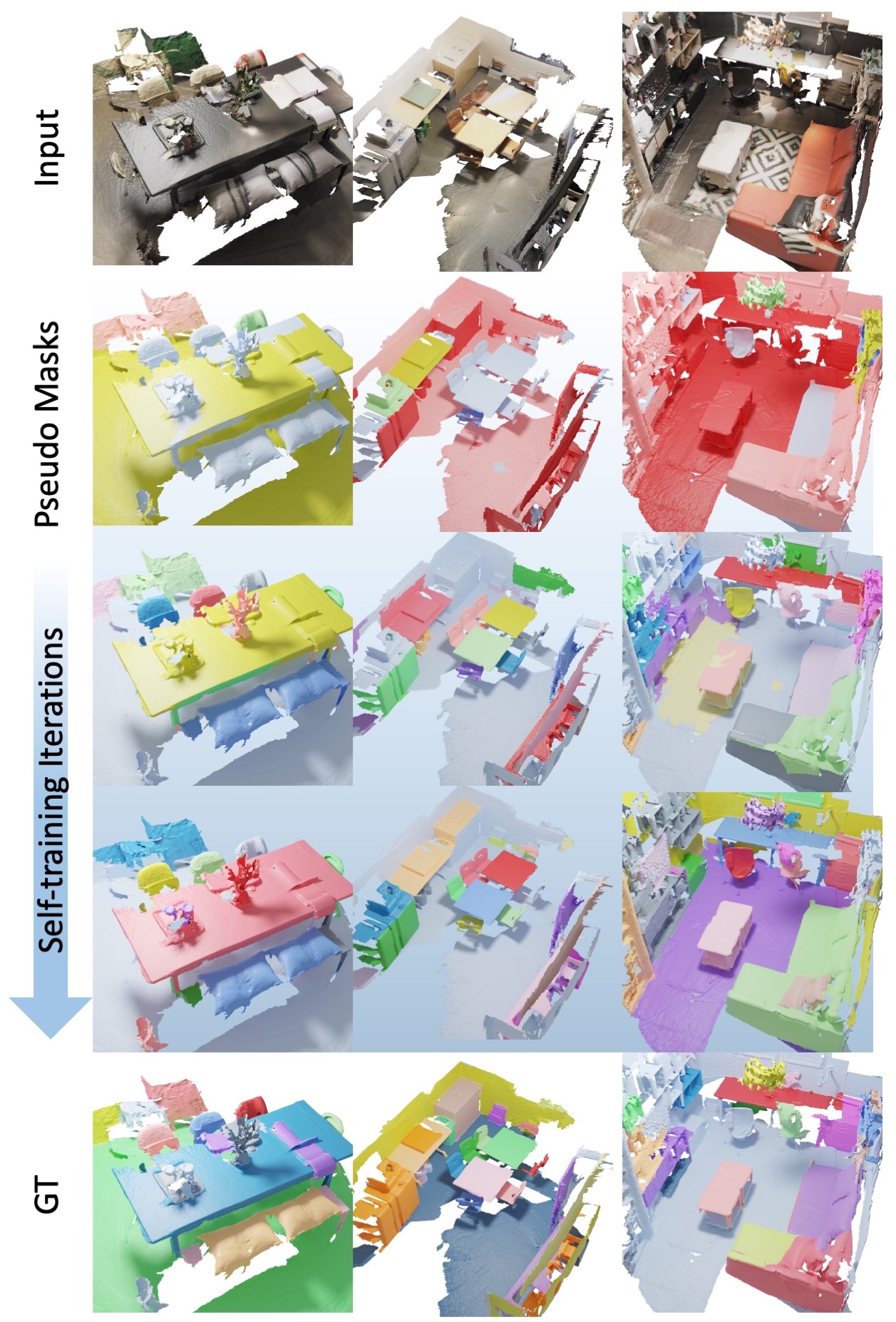}
    \caption[]{
    \OURS{} employs self-training to refine the initial sparse set of  proposals. We can see consistent improvement over both the number of predicted instances and the quality of the instance masks. Here we show results using the pseudo annotations obtained from both modalities. 
    } 
    \label{fig:self_training_refinement}
    \vspace{-0.4cm}
\end{figure}

%
After addressing the challenge of dimensionality reduction and effectively mitigating speckle noise in our features using geometric primitives, we can leverage the capabilities of the Normalized Cut algorithm to achieve a clean partitioning of scene graphs. For this, we iteratively apply NCut to our aggregated features for the extraction of initial pseudo masks denoted as $M$.
Starting with an empty set $M^0=\{\}$, we iteratively compute the adjacency matrix over $S_N$ and retrieve the masks $m \subset S_N$. 
We start from $N$ geometric segments with their corresponding $D$-dimensional features $\mathcal{F} \in \mathcal{R}^{N\times D}$, and construct the similarity matrix $A = sim(\mathcal{F})$, where $sim$ denotes cosine similarity. 
Additionally, for the multi-modal setup we calculate similarity matrices $A_{2D}$ and $A_{3D}$ independently and take their weighted average to obtain the final scores. 
Empirically, we found this to be more robust than direct feature fusion of the different modalities, due to their different statistical characteristics.
\indent We obtain $W_j$ introduced in Section \ref{sec:ncut_primitives} by thresholding $A$ at $\tau_{cut}$, where $j$ denotes the $j^{th}$ NCut iteration. 
Using $W_j$, we solve for the second eigenvector $v_j$ and threshold it to retrieve the partition $m_j$. 
We keep all separated foregrounds in $M^0$, where for each upcoming iteration, we mask out the row and column vectors from $W_j$, where $m_i \in M^0$ was already accepted as a foreground instance and $i$ being the previous segment ids.
This allows greedy separation of instances in order of confidence  in every cut iteration.
Examples of our generated pseudo masks are visualized in Figures \ref{fig:freemask_vs_ncut} and \ref{fig:self_training_refinement}. \\
\indent As the adjacency graph is unaware of the mesh connectivity, NCut often results in masks that span  spatially separated scene regions. 
In 3D, we can leverage knowledge of physical distance and connectivity of $G$ to constrain masks to be contiguous in the coarsened scene connectivity graph. 
We thus filter masks $m_j$ that have separated components, keeping only the parts $\tilde{m_j}$ that contain the item with the maximum absolute value in $v_j$. 
Separation based on connectivity is performed before saving $\tilde{m_j}$ into $M^0$, thus allowing for repeated detection of the dropped part over the next NCut iterations. 
Finally, we iterate until the maximum number of instances $M^0 = \{m_i\}_{i=1}^{N_m}$ are obtained, or there are no segments left in the scene. 
Moreover, we favor generating a reliable set of masks at the cost of restricting to a sparse initial set (i.e., missing potential instances rather than generating noisy masks for them) through a stricter $\tau_{cut}$ or lower number of instances.

\subsection{Self-Training}\label{sec:self_train}

Our initial pseudo masks can provide a set of proposed instances $M^0$; however, these pseudo masks are quite sparse in the scenes and sometimes over- or under-split nearby instances.
We thus refine the pseudo mask data through an iterative self-training strategy, producing final instance segmentation predictions $M'$ with more dense and complete instance proposals.

We leverage a state-of-the-art 3D transformer-based backbone~\cite{Schult23mask3d} for our self-training from pseudo mask data as mask-head supervision, while the class-head is collapsed to \textit{foreground} and \textit{background} classes.
Through multiple training cycles we save the proposals of the $t^{th}$ iteration into $M^{t}$, from the self-trained model, and save these masks as an extension to the original pseudo dataset obtaining $M^t \supseteq M^0$. 
From the second training iteration, we can extract the most confident $K$ predictions and sample these new instance proposals as an addition to the pseudo annotations. 
Further, we only accept new instances if the added information value is larger than the minimum threshold, measured by simple segment IoU scores. This way, we can effectively densify the originally sparse annotations, but without limiting the quality of the originally clean pseudo masks. 
\subsection{Implementation Details}\label{sec:implementation}
\paragraph{Backbones.} 
We use a Res16UNet34C sparse-voxel UNet implemented in the MinkowskiEngine~\cite{choy20194d} for 3D pre-trained feature extraction as well as for the 3D transformer during self-training. For the pretrained features we use our own trained weights of \cite{hou2021exploring} for compatibility reasons.

\paragraph{Self-training.} We employ the 3D transformer architecture of \cite{Schult23mask3d}, initialized from scratch. 
The first self-training cycle is trained for 600 epochs with a batch size of 8 until convergence, which takes $\approx 3$ days on a single NVIDIA RTX A6000 GPU. 
Further self-training cycles are all initialized from the previous state and finetuned for an additional 50 epochs in $\approx 4$ hours and for a total of 4 training cycles to produce the final set of instance predictions $S$. 
For the Hungarian assignment, we take the original weighted combination of dice and binary cross-entropy losses and only apply the DropLoss condition in the backpropagation phase.
\section{Experiments}

\begin{figure*}
    \centering
    \includegraphics[width=\linewidth,keepaspectratio]{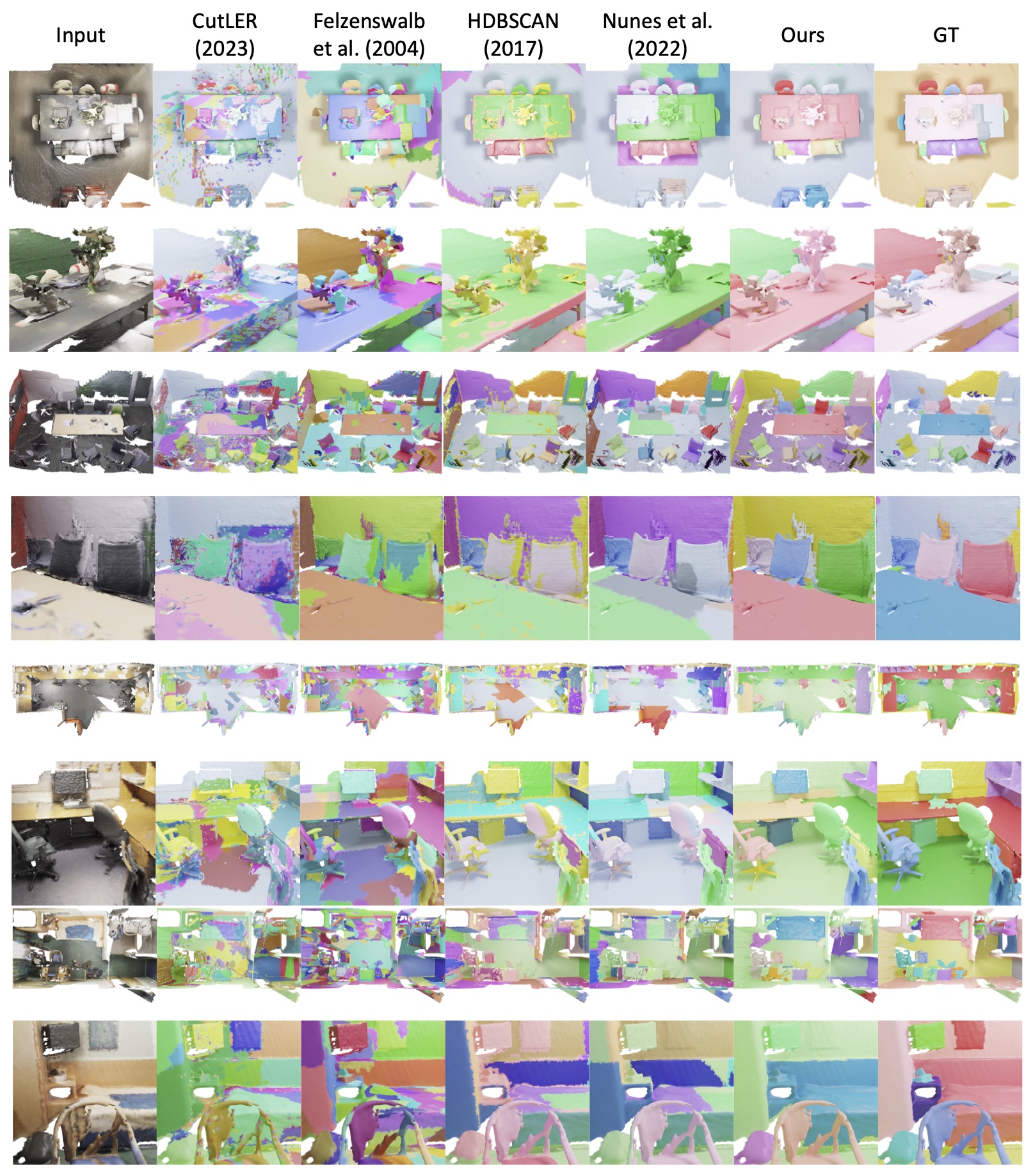}
    \caption[]{
    Qualitative comparison on ScanNet~\cite{dai2017scannet} scenes with projected predictions from the 2D method CutLER~\cite{wang2023cut}, traditional clustering-based methods Felzenszwalb~\cite{felzenszwalb2004efficient} and HDBSCAN~\cite{mcinnes2017accelerated_hdbscan}, and the GraphCut-based cluster refinement method \cite{nunes2022unsupervised}.
    Our approach leverages strong pseudo mask prediction and a self-training strategy to produce cleaner, more accurate instance segmentation.} 
    \label{fig:full_results}
\end{figure*}

We demonstrate the effectiveness of \OURS{} for unsupervised class-agnostic 3D instance segmentation on challenging real-world 3D scan datasets containing a large diversity of objects and significant clutter.
We train our method and all learned baselines on ScanNet~\cite{dai2017scannet}, using the official train split.
Note that no semantic annotation data is used for training, only the RGB-D reconstructions.
Additionally, we show that our approach trained on ScanNet data can effectively transfer to class-agnostic 3D instance segmentation on ARKitScenes~\cite{dehghan2021arkitscenes} data.

\vspace{-0.3cm}
\paragraph{Datasets.}
We train and evaluate \OURS{} on RGB-D scan data from ScanNet~\cite{dai2017scannet}, using the official train split.
We use the raw RGB images, and registered camera poses for training our approach, while the semantic annotations are used only for evaluation.
We use the official ScanNet train split for both the pre-trained 3D features from \cite{hou2021exploring} and our self-training iterations.
We additionally evaluate our method on ARKitScenes~\cite{dehghan2021arkitscenes}, on an 884/120 train/test split of indoor LIDAR scans. 
For ARKitScenes, we use 3D pre-trained features from ScanNet, followed by pseudo mask generation and self-training on the ARKitScenes train scenes.
We convert the LIDAR scan data to meshes with Poisson Surface Reconstruction~\cite{kazhdan2006poisson,kazhdan2013screened} prior to our graph coarsening.
Note that all baselines using learned features are trained on the same ScanNet data as ours.

\vspace{-0.3cm}
\paragraph{Evaluation metrics.}
We evaluate class-agnostic 3D instance segmentation performance with the widely-used Average Precision score on the full-resolution mesh vertices. Following the strategy of the supervised benchmark \cite{dai2017scannet} we report scores at IoU scores of 25\% and 50\% (AP@25, AP@50) and averaged over all overlaps between [50\% and 95\%] at 5\% steps (AP). 
Note that since predictions are class agnostic, all methods evaluate only instance mask AP values without considering any semantic class labels.
For ScanNet, we evaluate against ground truth instance masks from the established 20-class benchmark.
Since ARKitScenes does not contain any ground truth instance mask annotations, we evaluate all methods qualitatively.

\vspace{-0.3cm}
\paragraph{Comparison to the state of the art.}

\begin{table}
\centering
\begin{tabular}{lccccc}\toprule
  \textit{ScanNet}        & AP@25  & AP@50 & AP \\\midrule
HDBSCAN \cite{mcinnes2017accelerated_hdbscan}   & 32.1 & 5.5 & 1.6 \\
Nunes et al. \cite{nunes2022unsupervised}   & 30.5 & 7.3 & 2.3 \\
Felzenswalb \cite{felzenszwalb2004efficient}    & 38.9 & 12.7 & 5.0 \\
CutLER Projection \cite{wang2023cut}    & 7.0 & 0.2 & 0.3 \\
Ours    & \textbf{58.5} & \textbf{32.2} & \textbf{15.9} \\ \bottomrule
\end{tabular}
\caption{
Unsupervised class-agnostic 3D instance segmentation on ScanNet~\cite{dai2017scannet}. 
Our approach improves significantly over baselines (3x improvement in AP) due to our pseudo mask generation and self-training strategy.
}
\label{tab:scannet_results}
\vspace{-0.4cm}
\end{table}

We evaluate our approach in comparison to state-of-the-art traditional clustering methods HDBSCAN~\cite{mcinnes2017accelerated_hdbscan} and Felzenszwalb's algorithm~\cite{felzenszwalb2004efficient}, in addition to the unsupervised approach of Nunes et. al.~\cite{nunes2022unsupervised} leveraging learned feature clustering and refinement.
All baselines are provided with input mesh vertices, colors, and normals, while our approach and Nunes et. al. also operate on sparse voxel scene representations.
Table~\ref{tab:scannet_results} and Figure~\ref{fig:full_results} show comparisons on ScanNet data; our \OURS{} approach improves significantly over state of the art by effectively leveraging signal from self-supervised 3D features to guide our model through self-training.
Note that since Nunes et. al. has been designed for outdoor applications, even while leveraging ScanNet-trained features, it uses ground removal and relies on physical object separation, making segmentation difficult in cluttered scenes.

Additionally, we demonstrate the importance of reasoning in 3D, and compare with a state-of-the-art unsupervised 2D instance segmentation approach CutLER~\cite{wang2023cut} run on the RGB frames of the scans, and projected to 3D using the corresponding camera poses.
Here, the difficulty lies in resolving view inconsistencies, occlusions, and lack of knowledge of geometric structure resulting in poor 3D segmentation performance despite plausible 2D proposals.

\vspace{-0.3cm}
\paragraph{Evaluation on other datasets}
We quantitatively evaluate \OURS{} on the Area\_5 of the S3DIS dataset \cite{armeni_s3dis} using only 3D features pretrained on \cite{dai2017scannet}. Comparison with 3D-only state-of-the-art can be seen in Table \ref{tab:s3dis_results_table}.

\begin{table}[!ht]
\centering
\begin{tabular}{lccccc}\toprule
  \textit{S3DIS}        & AP@25  & AP@50 & AP \\\midrule
HDBSCAN \cite{mcinnes2017accelerated_hdbscan}   & 27.9   & 11.2  & 5.0 \\
Felzenswalb \cite{felzenszwalb2004efficient}    & 23.5   & 10.7  &   5.0  \\
Nunes et al. \cite{nunes2022unsupervised}   &  20.1  &  10.5 &  5.5 \\
Ours    &  \textbf{52.6}  &  \textbf{40.3} &  \textbf{21.4} \\ \bottomrule
\end{tabular}
\caption{Evaluation on S3DIS dataset (Area\_5). \OURS{} is able to adapt to other datasets as well and shows a significant improvement over previous SOTA methods.}
\label{tab:s3dis_results_table}
\vspace{-0.3cm}
\end{table}

\begin{figure}[!ht]
    \centering
    \includegraphics[width=\linewidth,keepaspectratio]{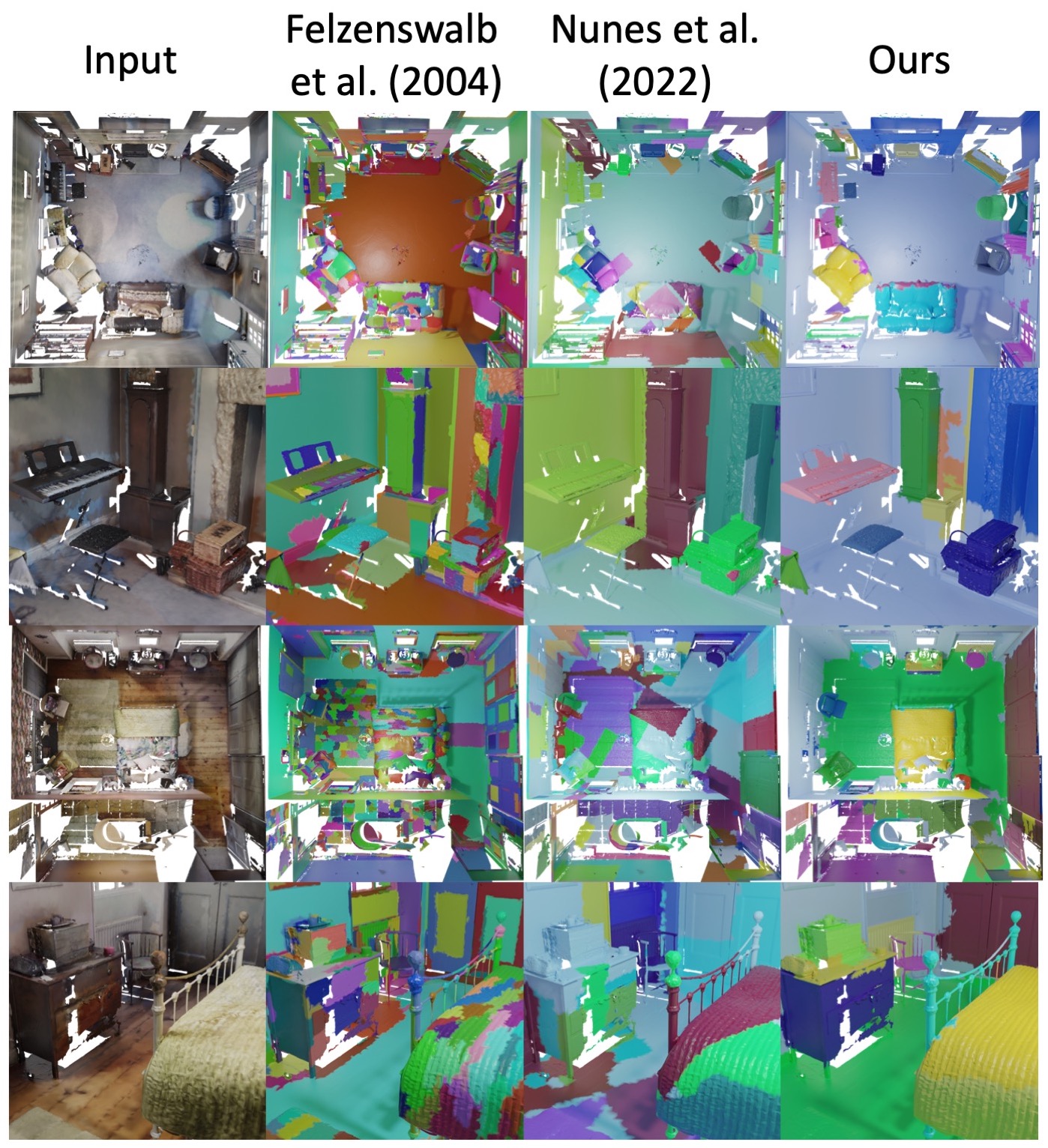}
    \caption[]{As \OURS{} does not require any human annotation, so we can also train and test our method on the ARKitScenes~\cite{dehghan2021arkitscenes} dataset. We leverages 3D features followed by a series of self-training iterations for cleaner, more accurate instance segmentation. Qualitative results shows consistently better results than our baselines. 
    } 
    \label{fig:arkitscenes_results}
    \vspace{-0.4cm}
\end{figure}

We additionally compare with state of the art on ARKitScenes~\cite{dehghan2021arkitscenes} data in Figure~\ref{fig:arkitscenes_results}. Here we show only qualitative results due to the absence of ground truth instance mask annotations. \OURS{} effectively produces cleaner, more accurate segmentations in these complex environments.

\begin{figure}[!ht]
    \centering
    \includegraphics[width=\linewidth,keepaspectratio]{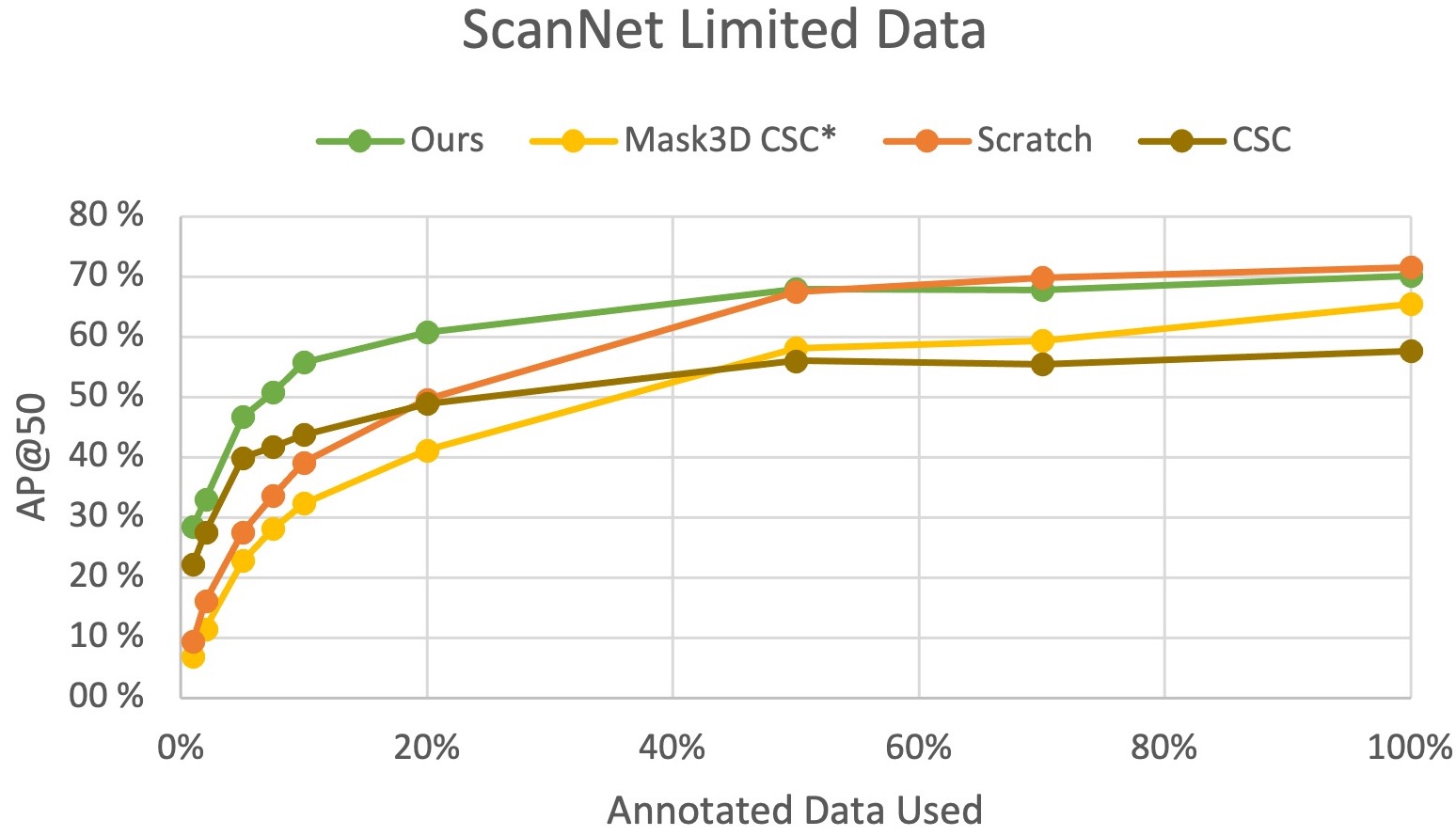}
    \caption[]{Our unsupervised self-training produces strong 3D features that can served as a powerful pretraining strategy for 3D instance segmentation in limited data scenarios. \OURS{} significantly outperforms state-of-the-art self-supervised 3D pretraining~\cite{hou2021exploring} on ScanNet instance segmentation. } 
    \label{fig:limited_data}
    \vspace{-0.3cm}
\end{figure}

\vspace{-0.3cm}
\paragraph{\OURS{} as data-efficient pretraining}
\OURS{} is able to learn powerful object properties and dense segmentation even in a fully unsupervised fashion. 
We demonstrate the potential of our strong learned features for downstream 3D instance segmentation with limited annotated data. 
We follow the setup introduced by CSC~\cite{hou2021exploring} with limited reconstructions available for downstream fine-tuning.
We show our method as a strong pretraining strategy in Figure~\ref{fig:limited_data}, notably outperforming both training from scratch as well as the state-of-the-art 3D pretraining of CSC. For more details we refer to our supplementary material.

\vspace{-0.3cm}
\paragraph{What is the effect of multi-modal signal for pseudo mask generation?} 
We evaluate the effect self-supervised color and geometry signals for generating pseudo annotations in Table~\ref{tab:pseudo_performance}. 
We consider using only self-supervised geometric features (3D), only self-supervised color features (2D) that are projected to the 3D scans, and both together (both).
We find that the color and geometry provide complementary signals.
We also note that color features are only used for the initial pseudo mask generation, during self-training iterations and test time only 3D features were used. 

\begin{figure}
    \centering
    \includegraphics[width=\linewidth,keepaspectratio]{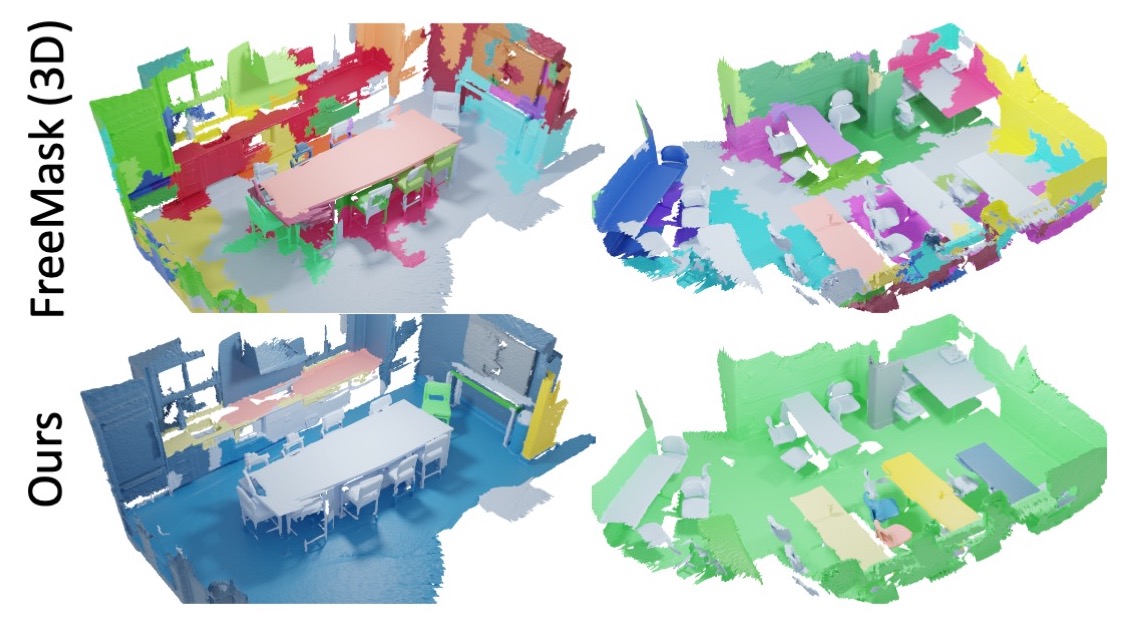}
    \caption[]{
    Initial pseudo masks generated by \OURS{} in comparison with a 3D-lifted FreeMask~\cite{wang2022freesolo}.
    FreeMask tends to produce a larger set of noisier pseudo masks, while we rely on a cleaner but sparser set for our self-training.
    } 
    \label{fig:freemask_vs_ncut}
\end{figure}

\vspace{-0.3cm}
\paragraph{What is the effect of pseudo annotations?}
We also evaluate the effect of our pseudo mask generation in  Table \ref{tab:pseudo_performance} and Figure \ref{fig:freemask_vs_ncut}, in comparison to the 3D adaptation of the FreeMask~\cite{wang2022freesolo} approach operating on our geometric segments. 
FreeMask tends to estimate a larger but noisier set of initial pseudo masks, while our approach is focusing on a sparser set of more reliable pseudo masks and produces significantly better performance. 
The strong difference in performance can be explained by the nature of the samples. While a sparser set of examples can be extended with multiple iterations of self-training, noisy samples will propagate through the full pipeline, and thus directly degrade the final performance. Further details of our adaptations of the FreeMask 3D method can be found in our supplemental.

\begin{table}
\centering
\small
\begin{tabular}{lcccccc}\toprule
           & Modality  & AP@25  & AP@50 & AP & AP Final\\\midrule
FreeMask   &  3D  &  14.4  & 3.6  &  1.3 &  2.0 \\
Ours    &  3D  &  45.4  &  16.7 &   9.2 &  13.3 \\  \midrule
FreeMask   &  2D  &  31.1  & 15.1 &  6.8 &  13.8 \\
Ours    &  2D  &  51.3  & 21.8 &   9.4  & 15.7 
\\ \midrule
FreeMask   &  both  &  23.7  &  10.1 &  5.7 & 12.1 \\
Ours    &  both  &  \textbf{52.9}  & \textbf{23.2}  &  \textbf{10.4} & \textbf{15.9} \\ \bottomrule
\end{tabular}
\caption{We compare pseudo mask generation from 3D-only features (3D), color-only features (2D), and both color and geometry (both) signal, as well as with pseudo annotation generation algorithm FreeMask~\cite{wang2022freesolo}. In this table we report method performances after a single iteration of self-training initialized from the different pseudo annotation methods and the final AP scores after 4 self-training iterations.}
\label{tab:pseudo_performance}
\vspace{-0.3cm}
\end{table}

\vspace{-0.3cm}
\paragraph{What is the impact of self-training?}
We observe that while self-training iterations are always improving the qualitative performance, their effective added information value is saturating after a limited number of cycles. We report on Table \ref{tab:self_train_iteration} through the first 4 steps, and observe a significant relative improvement in both modalities.  

\begin{table}
\centering
\resizebox{\linewidth}{!}{
\begin{tabular}{lcccccc}\toprule
& \multicolumn{3}{c}{3D Only} & \multicolumn{3}{c}{3D \& 2D}
\\\cmidrule(lr){2-4}\cmidrule(lr){5-7}
           & AP@25  & AP@50 & AP   & AP@25  & AP@50 & AP \\\midrule
$S^0$ pseudo masks  & 13.8 & 4.7  &  2 & 19.9 & 10.0 & 5.9 \\
$1^\textrm{st}$ Self-train  & 45.4 & 16.7  & 9.2  & 52.9 & 23.2 & 10.4 \\
$2^\textrm{nd}$ Self-train & 50.0 & 24.1  & 12.0  & 56.5 & 29.8 & 15.0 \\
$3^\textrm{rd}$ Self-train  & 52.2 &  25.8 & 12.8  & \textbf{58.8} & 31.9 & 15.9 \\
$4^\textrm{st}$ Self-train  & \textbf{52.7} &  \textbf{26.2} & \textbf{13.3}  & 58.5 & \textbf{32.2} & \textbf{15.9} \\ \bottomrule
\end{tabular}
}
\vspace{-0.3cm}
\caption{Multiple iterations of self-training significantly improve performance, saturating around 4 iterations.}
\label{tab:self_train_iteration}
\vspace{-0.3cm}
\end{table}

\vspace{-0.3cm}
\paragraph{Limitations}
While \OURS{} offers a promising step towards unsupervised 3D instance segmentation, various limitations remain.
We rely on a mesh representation for graph coarsening, but believe this could be extended to alternative representations through neighborhood reasoning.
Additionally, our graph coarsening step may cause very small objects (e.g., pens, cell phones) to be missed in the pseudo annotation generation.
Finally, employing a fixed set of pseudo masks from the initial stage that are used through self-training could reinforce noisy predictions.

\section{Conclusion}
We introduced \OURS{}, a novel approach towards achieving fully-unsupervised 3D instance segmentation in cluttered indoor scenes. 
Our approach effectively combined low-level geometric properties to regularize multi-modal self-supervised deep features for initial pseudo mask extraction, and our self-training notably improved performance by refining these proposals to a more complete, dense set of instances.
As 3D instance segmentation is a crucial aspect of 3D scene understanding, \OURS{}'s  ability to achieve this without requiring any manual annotations opens up new possibilities for 3D semantic understanding.

\section{Acknowledgements}
    This project is funded by the Bavarian State Ministry of Science and the Arts and coordinated by the Bavarian Research Institute for Digital Transformation (bidt), the ERC Starting Grant SpatialSem (101076253), and supported in part by a Google research gift. Or Litany is a Taub fellow and is supported by the Azrieli Foundation Early Career Faculty Fellowship.

{
    \small
    \bibliographystyle{ieeenat_fullname}
    \bibliography{main}
}

\clearpage
\section{Appendix}

%
\begin{figure*}[!ht]
    \centering
    \includegraphics[width=0.75\linewidth, keepaspectratio]{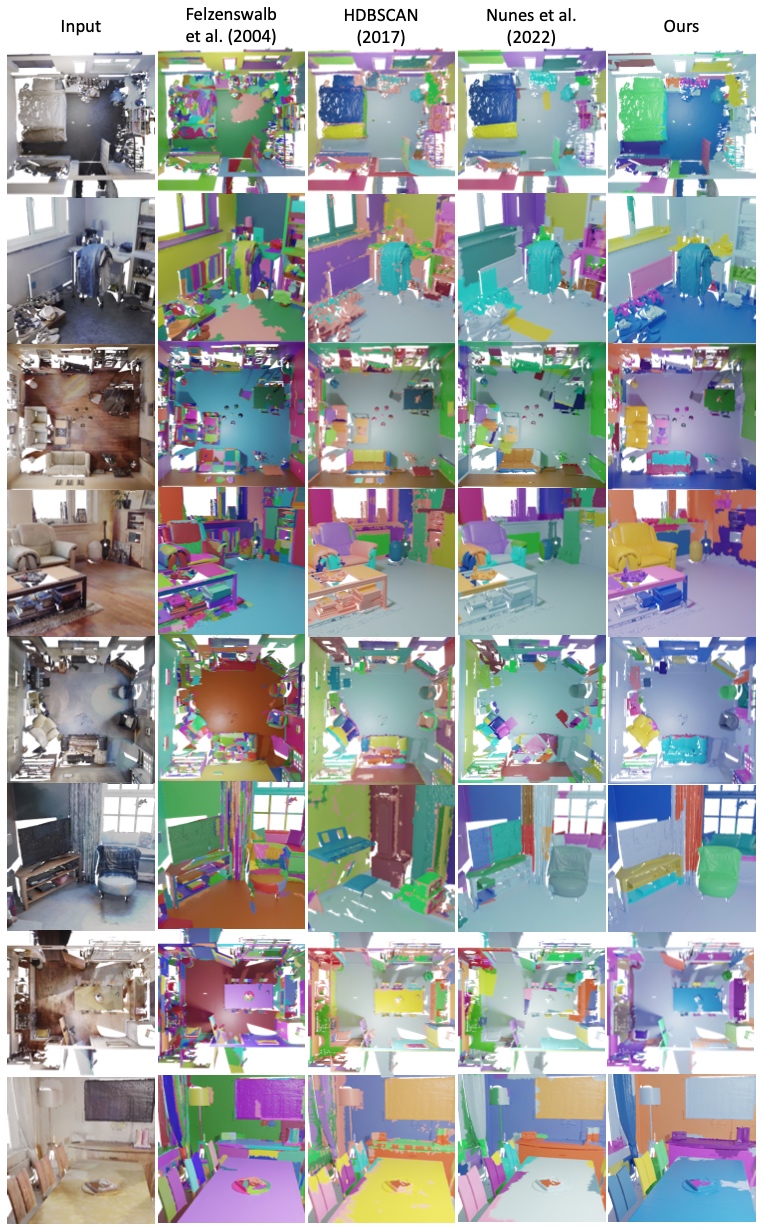}
    \caption[]{Additional results on the ARKitScenes dataset \cite{dehghan2021arkitscenes}, compared to geometric clustering and oversegmentation-based baselines.} 
    \label{fig:arkit_supplement}
\end{figure*}
\begin{figure*}[!ht]
    \centering
    \includegraphics[width=0.85\linewidth,keepaspectratio]{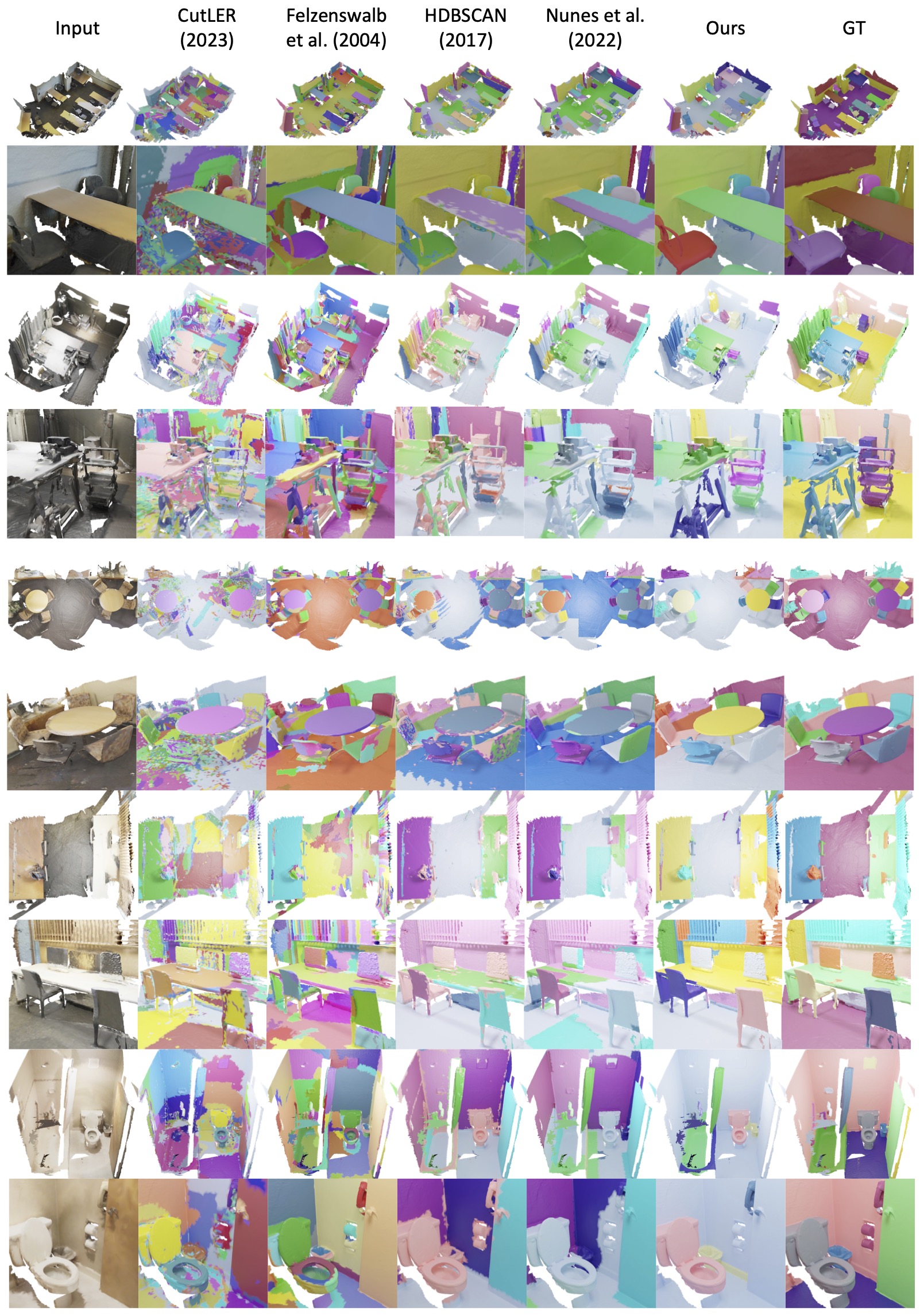}
    \caption[]{Additional results on the ScanNet dataset \cite{dai2017scannet}, compared to geometric clustering and oversegmentation-based baselines.} 
    \label{fig:scannet_supplemental}
\end{figure*}

\subsection{\OURS{} as Data Efficient Pretraining}

We report additional qualitative details on the data efficient pretraining performance of \OURS{} in Table~\ref{tab:data_efficient}.

We also note that the 3D contrastive pre-training of CSC, similar to other 3D pre-training methods developed for non-transformer backbones \cite{xie2020pointcontrast, hou2021exploring, zhang_depth_contrast,nunes2022segcontrast}, was not beneficial for a transformer-based model. A similar observation was also reported in a recent pretraining method \cite{hou2023mask3d}. 
We thus also compare with CSC pretraining on their original 3D backbone (which demonstrated improvement over training from scratch on the same backbone).
Our approach can improves notably over both alternatives.
\begin{table*}[!ht]
\centering
\resizebox{\linewidth}{!}{
\begin{tabular}{ccccccccccccccccc}\toprule
& & \multicolumn{3}{c}{1\%} &  \multicolumn{3}{c}{5\%} &  \multicolumn{3}{c}{10\%} &  \multicolumn{3}{c}{20\%} & \multicolumn{3}{c}{50\%}
\\\cmidrule(lr){3-5}\cmidrule(lr){6-8}\cmidrule(lr){9-11}\cmidrule(lr){12-14}\cmidrule(lr){15-17}
  Model & Backbone  & AP@25  & AP@50 & AP   & AP@25  & AP@50 & AP & AP@25  & AP@50 & AP & AP@25  & AP@50 & AP & AP@25  & AP@50 & AP \\\midrule
Scratch & Bottom-up & 22.6 &14.1 &6.8 &45.5 &33.3 &18.1 &54.8 &39.2 &21.9 & 61.0 &43.4 &25.5 &67.0 &51.4 &30.3 \\
CSC \cite{hou2021exploring} & Bottom-up &35.6 &22.1 &12.5 &52.7 &39.9 &23.3 &59.8 & 43.8& 25.0& 63.8& 48.9& 29.6& 70.5& 56.0 & 33.6 \\
Scratch & Transformer & 24.7 & 9.3 & 4.6 & 48.1 & 27.6 & 16.3 & 59.2 & 39.1& 23.4& 66.4 & 49.6 & 33.1& \textbf{78.9}& 67.5& \textbf{49.8} \\
CSC  & Transformer & 17.0& 6.8& 3.8& 44.2& 22.7& 13.1& 55.2& 32.3& 19.1& 62.0& 41.2& 26.0& 73.7& 58.2&40.0 \\
Ours & Transformer & \textbf{43.5} & \textbf{28.4} & \textbf{15.8} & \textbf{63.2} & \textbf{46.8} & \textbf{28.3} & \textbf{70.3} & \textbf{55.7} & \textbf{36.7} & \textbf{72.4} & \textbf{60.7}& \textbf{41.5} & \textbf{78.9}& \textbf{68.0} & 48.2 \\ \bottomrule
\end{tabular}
}
\caption{Unsupervised class-agnostic pretraining with our method can also act as a powerful pretraining strategy, advancing over state of the art. 
We report pretraining with CSC \cite{hou2021exploring} and \OURS{}, and evaluate the downstream weakly-supervised instance segmentation performance on ScanNet with percentage of limited annoated scenes used denoted in the top row. 
As we found that CSC degraded performance when using a transformer-based backbone, we also report the performance of training from scratch and CSC on their originally proposed backbone of a sparse UNet with bottom-up voting.}
\label{tab:data_efficient}
\end{table*}

\subsection{The effect of noise robust losses.} 
We adopt DropLoss \cite{wang2023cut} for our self-training cycles, which is robust to sparse data and missing annotations. 
In particular, we use a weighted combination of cross-entropy and Dice \cite{sudre2017generalised_diceloss} losses for bipartite-matching with pseudo annotations.
We then drop losses for backpropagation which do not have at least $\tau_{drop}$ overlap with the annotations from the previous cycle.
We evaluate the effect of different noise robust losses for self-training in Table~\ref{tab:noise_robust_ablation}. 
We compare our baseline losses with a 3D extension of the projection loss of \cite{wang2022freesolo}, and our adaptation of  DropLoss from \cite{wang2023cut}.
Our approach does not penalize for missing pseudo masks, which enables more effective self-training to discover previously missed instances.

\begin{table}[!ht]
\centering
\small
\begin{tabular}{lcccc}\toprule
     & AP@25  & AP@50 & AP & AP Final\\\midrule
Initial Pseudo Masks &  19.9  &  10.0  & 5.9  & -\\
Baseline losses \cite{Schult23mask3d} &  42.3  &  16.9 &  7.2 &  14.2 \\
Projection loss \cite{wang2022freesolo} &  35.7  &  12.1  &  4.7 & 7.2 \\
DropLoss \cite{wang2023cut} &  \textbf{52.9}  &  \textbf{23.2} &  \textbf{10.4}  & \textbf{15.9} \\ \bottomrule
\end{tabular}
\caption{
A 3D projection loss struggles with under-determined associations, while DropLoss helps \OURS{} to discover parts of the scene that were missed by the source supervision. We report all metrics after a single iteration and the AP scores after 4 iterations of self-training.}
\label{tab:noise_robust_ablation}
\end{table}

\subsection{Additional Qualitative Results}

We show more qualitative results from our method trained on ARKitScenes \cite{dehghan2021arkitscenes} in Figure \ref{fig:arkit_supplement} and on ScanNet \cite{dai2017scannet} in Figure \ref{fig:scannet_supplemental}. 

\subsection{Pseudo Mask Generation Ablations}

We also ablate the saliency threshold, oversegmentation parameters, and separation strategy  in our pseudo mask generation. If not explicitly stated otherwise in Table \ref{tab:freemask_vs_ncut}, we use both 2D and 3D modality features for the pseudo mask generation. 

\paragraph{What is the effect of the saliency threshold in pseudo mask generation?} 

 We threshold the saliency matrix $A$ with $\tau_{cut}=0.55$ for geometric-only features and $\tau_{cut}=0.65$ for combined modalities. Table~\ref{tab:cutler3d_tau} shows that our approach maintains robust performance across a large range of $\tau_{cut}$ thresholds used to estimate salient areas for pseudo masks. In this table we report results using features from combined modalities, but similar behaviour can be observed for the other scenarios as well.

\begin{table}
\centering
\small
\begin{tabular}{lccc}\toprule
$\tau_{cut}$ & AP@25  & AP@50 & AP \\\midrule
0.40 & 16.7 & 9.0 & 5.2 \\
0.50 & 20.8 & 10.7 & 5.7 \\
0.55 & 21.0 & 10.8 & 5.7 \\
0.60 & 21.3 & 11.3 & 5.8 \\
\textbf{0.65} & \textbf{19.9} & \textbf{10.0} & \textbf{5.9} \\
0.70 & 18.2 & 9.9 & 5.6 \\
0.80 & 11.8 & 5.0 & 2.6 \\ \bottomrule
\end{tabular}
\caption{
Our pseudo mask generation quality, as measured by AP metrics, maintains robustness to a large range of $\tau$ thresholds that extract saliency.
Note that this measures the quality of only the pseudo masks; our full approach with self-training produces significantly improved results. In this table we show results and parameters used by our method in bold and report pseudo mask performance generated from both modalities.}
\label{tab:cutler3d_tau}
\end{table}

\paragraph{The effect of iterative mask densification.}

We designed a strategy to leverage a sparse set of relatively clean initial pseudo masks, which are progressively extended with confident self-predictions during later iterations. This leads to a 3x improvement over state of the art in the Average Precision Metric. We could also consider different mask refinement strategies using a mixture of segments, initial masks or self-trained instances.
Tab.~\ref{tab:mask_refinement} ablates a mask refinement strategy of discarding previous masks and retaining current predictions. We also consider using Felzenswalb segments directly instead of feature-based pseudo labels. Both these strategies lead to lower performance due to the increased presence of noisy labels, which dominate the training signal.

\vspace{-0.3cm}
\begin{table}[!ht]
\centering
\footnotesize
\begin{tabular}{lccc}
                               & AP@25 & AP@50 & AP    \\ \hline \hline
\multicolumn{1}{l|}{Felzenswalb Masks}   & 35.5   & 20.6  &  10.3  \\ \hline
\multicolumn{1}{l|}{Mask Refinement}   &   43.7 & 24.4  &  12.4  \\ \hline
\multicolumn{1}{l|}{Mask Addition (Ours)}   &  58.6  & 32.0  &  16.0  \\ \hline
\end{tabular}
\caption{Instead of using masks from previous iteration directly it is the best to keep the initial masks fixed, and iteratively sample plausible predictions to enrich the pseudo dataset during self-training. This method strikes a balance between relatively clean, but sparse labels and increasing number of confident samples. Finally, even though Felzenswalb oversegmentation yields to higher precision, then our initial mask prediction algorithm, it also includes more background into the training, and this way plateauing at a lower self-training performance.}
\vspace{-0.3cm}
\label{tab:mask_refinement}
\end{table}

\paragraph{Robustness to oversegmentation parameters.}
Table~\ref{tab:pseudomaskablation} shows that our approach maintains strong robustness to a wide range of oversegmentation parameters for our geometric segments (our used parameters denoted in bold).

\paragraph{Additional pseudo mask generation hyperparameters.} 
Additionally, we also test the effect of other hyperparameters in out \textit{NCut}-based pseudo mask generation module, including  used distance metrics in the similarity matrix and different methods to separate unconnected patches in the predicted foregrounds. 
During the foreground separation in the Normalized Cut algorithm, we had an additional condition for the minimum number of foreground segments for the bipartitions. This conditions was able effectively filter out suboptimal partitioning of the full graph leading to separated parts from the full instances. Reducing the size of this parameter can directly lead to a more dense set of initial pseudo masks, with the cost of higher false positive rate. In Table \ref{tab:pseudomaskablation} we report a sparser and denser version of the datasets with a minimum number of foregorund segments of 8 and 2 accordingly, and show the initial higher scores of the pseudo annotation doesn't necessarily propagate to better downstream self-trained performance. 

Finally, we also ablate the effect of our physical connectivity-based foreground separation introduced in Section 3.1. In our main method we separate all set of connected components in the foreground, but only keep the component with the highest eigenvector activation (\textit{Max}). As an alternative we also test a method where we calculate the highest average activation in the connected component (\textit{Avg.}), a method where we keep the component with the largest surface value (\textit{Largest}) and finally, to test the effect of this module, without any kind of connectivity-based separation (\textit{No Sep.}). 

\begin{table*}
\centering
\resizebox{\linewidth}{!}{
\begin{tabular}{ccccccccccccccc}\toprule
\multicolumn{4}{c}{Generation Params.} & \multicolumn{4}{c}{Initial Pseudo Mask} & \multicolumn{3}{c}{1 Iteration of Self-Training}  & \multicolumn{3}{c}{4 Iterations of Self-Training}
\\\cmidrule(lr){1-4}\cmidrule(lr){5-8}\cmidrule(lr){9-11}\cmidrule(lr){12-14}
  Segment Size & Metric  & Separation & Min. \# of Foreground & \# of Instances & AP@25  & AP@50 & AP  & AP@25  & AP@50 & AP & AP@25  & AP@50 & AP  \\\midrule
30 & Cos & Max & 8 & 2169 &  21.9 & 11.5 & 6.3  & 53.7 & 26.2 & 12.4 & 55.4 & 30.3 & 15.3\\
 \textbf{50} & \textbf{Cos} & \textbf{Max} & \textbf{8} & 1414 & \textbf{19.9} & \textbf{10.0} & \textbf{5.9}  & \textbf{52.9} & \textbf{23.2} & \textbf{10.4} & \textbf{58.5} & \textbf{32.2} & \textbf{15.9} \\
 100 & Cos & Max & 8 & 1090 &  17.4 & 8.0 & 4.2 & 33.1 & 10.2 & 3.9 & 39.6 & 13.7 & 5.3 \\
 200 & Cos & Max & 8 & 584 &  11.0 & 3.7 & 1.8  & 24.3 & 8.7 & 2.1 & 26.1 & 9.7 & 2.4\\
 400 & Cos & Max & 8 & 319  & 6.4 & 2.5 & 1.1  & 19.1 & 3.9 & 1.2 & 19.9 & 3.2 & 1.0 \\ \midrule
 50 & L2 & Max & 8 & 1539 &  20.1 & 10.6 & 5.4 &  49.0 & 21.7  & 9.8 & 55.3 & 38.4 & 14.3 \\
 100 & L2 & Max & 8 & 805 & 13.3 & 5.3 & 2.6 & 30.8 & 8.3 & 2.8  & 39.0 & 12.7 & 5.0\\ \midrule
 50 & Cos & No Sep. & 8 & 125 &  4.3 & 0.3 &  0.1  & 4.3 & 0.5 & 0.2 & 4.9 & 0.6 & 0.2 \\ 
 50 & Cos & Largest & 8 & 620 &  11.5 & 4.9 & 2.5 &  11.5 & 1.5 & 0.4 & 12.9 & 2.2 & 12.9\\
 50 & Cos & Avg.  & 8 & 1078 & 16.8 & 9.1 & 5.1  & 36.4 & 12.5 & 4.9 & 43.8 & 17.8 & 7.5 \\ \midrule \midrule
 30 & Cos & Max & 2 & 2909 & 29.0 & 15.6 & 8.7 & 53.6 & 28.6 & 14.2 & 54.2 & 29.8& 15.4\\
 50 & Cos & Max & 2 & 2512 &24.9 & 12.4 & 7.2 & 56.5 & 29.8 & 15.0  & 51.3 & 26.2& 12.6\\
 100 & Cos & Max & 2 & 2317 & 23.1 & 12.3 & 6.8 & 51.8 & 24.4 & 11.6  & 57.1 & 31.3& 15.6\\
 200 & Cos & Max & 2 & 2181 & 28.4 & 15.5 & 8.9 & 54.6 & 28.7 &  13.7 & 56.6&31.4 & 15.6\\
 400 & Cos & Max & 2 & 1373 & 20.6 & 11.1 & 6.3 & 51.0 & 24.8 &  11.8 & 55.8 & 30.3& 15.2\\ \midrule
 50 & L2 & Max & 2 & 2496 & 28.6 & 15.8 & 9.0 & 55.8  & 29.6 & 14.6 & 54.8& 30.3& 15.3\\
 100 & L2 & Max & 2 & 1668 & 23.4 & 12.7 & 7.3 & 53.1 & 25.0 & 11.3 & 56.3&27.7 &12.9 \\ \midrule
 50 & Cos & No Sep. & 2 & 159 & 0.2 & 0.5 & 3.6 &  5.4 & 0.6 & 0.3 & 3.9& 0.4& 0.2\\ 
 50 & Cos & Largest & 2 &1026 & 14.1 & 7.2 & 3.9 & 11.5 & 1.8 & 0.5  &14.5 & 2.5& 0.7\\
 50 & Cos & Avg.  & 2 & 2053 & 23.3 & 12.0 & 6.8 & 52.5 & 27.4 & 12.7 & 54.9& 29.9& 14.9\\ \bottomrule
\end{tabular}
}
\caption{
We denote the parameters used by our method in bold. 
We show that our method is robust to a wide range of numbers regarding segments sizes and different similarity metrics, and only degrades somewhat in performance when segments are constrained to be too large. 
We also show that the separation of physically distant foreground patches is important and it is beneficial to use the activation of the eigenvector for the best results. 
Finally, we show that denser initial mask predictions lead to quantitatively better initial pseudo annotations, and even better self-training performance after a single iteration, but underperforming in their final scores. This behaviour can be explained by the larger false positive ratio in the denser initial predictions, which is propagating through all iterations, but thanks to the noise robust losses and iterative refinement of predictions the sparse set of labels can be effectively used. In this table we report results using both modalities for the initial pseudo mask generation, and number predicted pseudo instances in the official validation split of the ScanNet dataset.}
\label{tab:pseudomaskablation}
\end{table*}

\subsection{Comparison with methods from the 2D domain}

To ensure a fair evaluation of methods operating on different input domains in Table 1. we followed the established procedure of well-known baselines \cite{Dai20183DMVJ3,hou20193d,jaritz2019multi}. This involves using depth information to project 2D predictions into 3D such that all methods are evaluated in the same 3D domain and aggregate multiple predictions through consensus by majority voting or accepting the maximum confidence scores for every voxel location. 
We also show results evaluated against 2D ScanNet images by projecting our method's predictions into 2D in Tab.~\ref{tab:evaluation_2d}, and comparing it to the current state of the art 2D unsupervised segmentation method \cite{wang2023cut} which demonstrates the usefulness of 3D reasoning.

\vspace{-0.3cm}s
\begin{table}[!ht]
\centering
\resizebox{\columnwidth}{!}{
\begin{tabular}{lccc}
                               & AP@25 (2D) & AP@50 (2D) & AP (2D)    \\ \hline \hline
\multicolumn{1}{l|}{CutLER (2D)}   &  7.8  & 2.8  &  0.7  \\ \hline
\multicolumn{1}{l|}{Ours (projected)}   &  60.0  & 38.1  &  21.1  \\ \hline
\end{tabular}}
\caption{2D evaluation on ScanNet images.}
\vspace{-0.3cm}
\label{tab:evaluation_2d}
\end{table}

\begin{figure}[!h]
    \centering
    \centering
    \includegraphics[width=\linewidth]{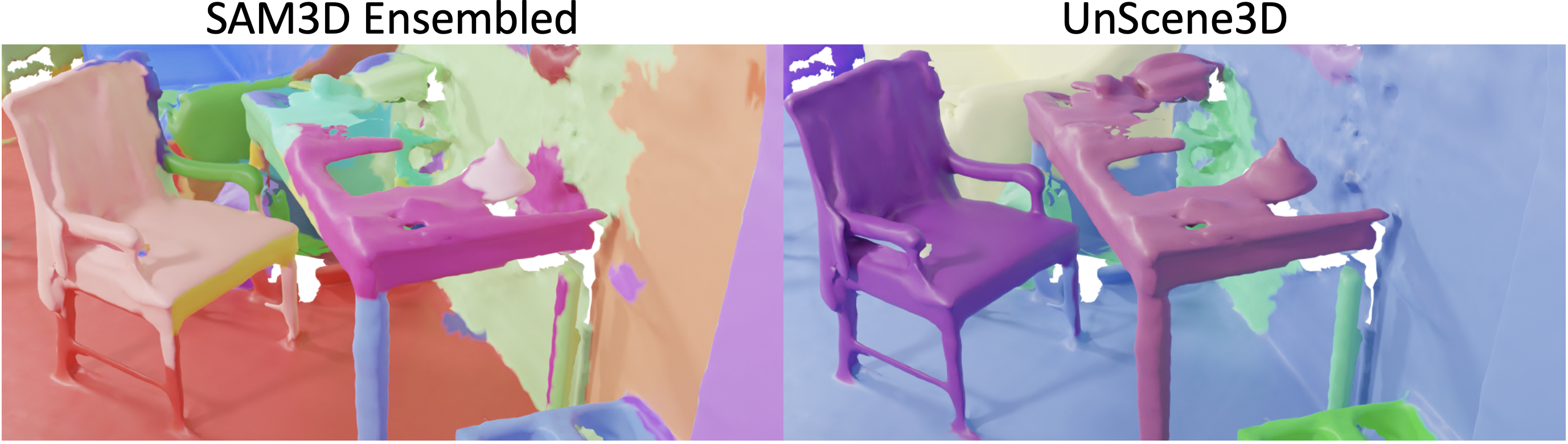}
    \vspace{-0.8cm}
    \caption{ While SAM has powerful capabilities in crisp 2D mask generation, when aggregated on 3D, SAM3D tends to over-segment object instances. }
    \label{fig:sam3d_preds}
    \vspace{-0.5cm}
\end{figure}

We also compare to weakly-supervised instance segmentation method SAM3D \cite{yang2023sam3d}, where powerful class-agnostic 2D masks are extracted by the powerful SAM model \cite{kirillov2023segment}. Here the projected 2D masks are merged into 3D masks iteratively with a bottom-up bidirectional merging approach to achieved cleaner and more view-independent 3D instances. A qualitative comparison on ScanNet can be seen in Table \ref{tab:sam3d_results}, with qualitative comparisons in Figure \ref{fig:sam3d_preds}.

\vspace{-0.3cm}
\begin{table}[!h]
\centering
\footnotesize
\begin{tabular}{lccc}
                               & AP@25 & AP@50 & AP    \\ \hline \hline
\multicolumn{1}{l|}{SAM3D}   & 37.2   &  11.8 &   3.7 \\ \hline
\multicolumn{1}{l|}{SAM3D with GT Segments}   & 47.6   & 24.1  & 10.8   \\ \hline
\multicolumn{1}{l|}{Ours}   & \textbf{58.5}   & \textbf{32.2}  &  \textbf{15.9}  \\ \hline
\end{tabular}
\vspace{-0.3cm}
\caption{\OURS{} achieves significantly better performance on ScanNet than SAM3D through our strong multi-modal reasoning. 
}
\vspace{-0.7cm}
\label{tab:sam3d_results}
\end{table}

SAM3D must resolve view inconsistencies and SAM’s tendency to over-segment objects, which results in SAM3D splitting instances, while \OURS{} is able to achieve complete masks through multi-modal reasoning.
We believe integrating SAM or other (weakly-) supervised 2D models into our pipeline to enable multi-modal reasoning is an interesting avenue for future work.

\subsection{Additional Implementation Details}

Here, we further explain the implementation details of our pseudo mask generation. 

\paragraph{Pseudo code for masked NCut}

We show the pseudo code-style implementation for the masked normalized cut algorithm generating multiple instances as pseudo masks. The full algorithm can be seen in \ref{alg:masked_ncut}. 

\begin{algorithm}[ht!]
\DontPrintSemicolon
\caption{Masked NCut on 3D segments} \label{alg:masked_ncut}
\KwData{$\mathcal{S} = \{s_i,  \dots, s_N\}$, $\mathcal{F} \in \mathcal{R}^{NxD}$, 
$\mathcal{C} = \{(s_1, s_k), (s_1, s_l), \dots \}$}
\KwResult{$\mathcal{M} = \{m_j,  \dots, m_M\}$}
$\mathcal{M} \gets \{\}$ \\
\While{$j \le max\_inst\_num$}{
  $\mathcal{F}' \gets \mathcal{F}$ \\
  $\mathcal{F}'[\mathcal{M}] \gets 0.$ \tcp*{Mask out previous insts.}
  $\mathcal{W} \gets \mathcal{F} \times \mathcal{F}^T$ \tcp*{Feature similarity}
  \tcp*[l]{Saliency with connected graph}
  $\mathcal{W}_{i,k} = \left\{\begin{array}{cl}
           1. & \text{if $\mathcal{W}_{i,k} \geq \tau_{cut}$} \\
           \epsilon               & \text{if $\mathcal{W}_{i,k} < \tau_{cut}$} \\
           \end{array}\right.$ \\  \label{algoline:saliency}
  $\mathcal{D}_{i,i} = \sum_{k} W_{i,k}$ \\
  \tcp*[l]{Get $2^{nd}$ smallest eigenvector}
  $\lambda, \mathbf{v} \gets eigh(\mathcal{D} - \mathcal{W}, \mathcal{D}, -2)$ \\
  $m_i = \left\{\begin{array}{cl}
           1 & \text{if $ v_i \geq mean( \mathbf{v} )$} \\
           0 & \text{if $ v_i < mean( \mathbf{v} )$} \\
           \end{array}\right.$ \\   \label{algoline:segment_foreground}
  \tcp*[l]{Invert bipartition if too large}
  \If{$sum(\mathbf{m}) > D / 2$}{
    $ \mathbf{m} = 1 - \mathbf{m} $ \\
    $ \mathbf{v} = -1. * \mathbf{v} $ \\ 
    }
  \tcp*[l]{Separate unconnected components}
  $ v_{max} = max(\mathbf{v}) $ \\
  $ \tilde{\mathbf{m}} = sep(\mathbf{v}, v_{max}, \mathcal{C})$ \\
  $M \gets M \cup \{\tilde{\mathbf{m}}\}$
}
\end{algorithm}

\paragraph{3D Adaptation of FreeMask}

We also evaluate an alternative pseudo mask segmentation algorithm besides the masked \textit{NCut} method. 
In the 2D domain FreeSOLO \cite{wang2022freesolo} also followed a two stage pipeline first generating the pseudo annotations, and then refine those predictions through a series of self-training cycles. We followed their intuition to take a self-supervised pretrained backbone and extract it's deep features at multiple levels of the decoder. 
While in standard pretrained UNet-style models early features represent global context, final features and local semantic meaning, intermediate features can act as an useful proxy to extract self-similar regions in the input samples. 
In our implementation we used the same backbone features of \cite{hou2021exploring,caron2021emerging_dino} for the same 2D-3D setup and extracted the penultimate layer features for the self-similarity calculation. 
Then sampled the feature space with the Furthest Point Sampling \cite{qi2017pointnet++} strategy to get a more limited set of anchor points, later used to extract self-similar regions. 
For every seed point we took similarity scores with the other features of the full scene and thresholded it to extract salient regions. 
Finally, we used the efficient Non Maximum Suppression implementation from \cite{wang2022freesolo} to sort the predicted salient areas and filter out overlapping regions. 
We also used average similarity score combined with the salient region area to get \textit{maskness scores} for every salient region, directly following the original implementation. 
We report comparative results of the masked \textit{NCut} algorithm and our FreeMask 3D adaptation after self-training in Table 3. of the main paper and in Table \ref{tab:freemask_vs_ncut} of the initial pseudo mask scores. 

\begin{table}[!ht]
\centering
\small
\begin{tabular}{lccccc}\toprule
           & Modality  & AP@25  & AP@50 & AP \\\midrule
FreeMask   &  3D  &  13.7  & 7,2  &  3.7   \\
Ours    &  3D & 13.8 & 4.7 & 2.0  \\  \midrule
FreeMask   &  2D  & 15.3 & 6.6 & 2.9  \\
Ours    &  2D  & 15.6 & 7.2 & 3.6   \\ \midrule
FreeMask   &  both  & 17.9 & 7.5 & 3.7  \\
Ours    &  both  & 19.9 & 10.0 &  5.9 \\ \bottomrule
\end{tabular}
\caption{We compare pseudo mask generation from 3D-only features (3D), color-only features (2D), and both color and geometry (both) signal, as well as with pseudo annotation generation algorithm FreeMask.
We compare the quality of the initial pseudo mask dataset using our masked \textit{NCut} algorithm and the adaptation of FreeMask~\cite{wang2022freesolo} to 3D. We see that the normalized cut-based method is superior for both modalities.}
\label{tab:freemask_vs_ncut}
\end{table}

We also note here that while there is a difference in the initial pseudo mask qualities for the different methods, the downstream performance is way more significant. This can explained by the nature of the pseudo masks. \textit{NCut} provides a clean and sparse set of annotation, which is easy to densify for following iterations. On the other hand, the more dense, but noisy FreeMask predictions remain in the training for the duration of the whole training, hindering the performance of the self-trained model with noisy supervision.


\end{document}